\documentclass[letterpaper, 10 pt, conference]{ieeeconf}  %

\IEEEoverridecommandlockouts                              %

\usepackage{soul}
\usepackage{color}
\usepackage{wallpaper}
\usepackage{graphicx}
\usepackage{cleveref}
\usepackage{numprint}
\usepackage{graphicx}
\usepackage{subcaption}
\usepackage{mwe}
\usepackage{multirow}
\usepackage{makecell}

\definecolor{orangeX}{rgb}{1,.5,0}
\definecolor{blueX}{rgb}{.2, .59, .88}
\definecolor{purpleX}{rgb}{.294118, 0, .509804}
\definecolor{greenX}{rgb}{.421, .578, .241}
\definecolor{bole}{rgb}{0.47, 0.27, 0.23}
\definecolor{mypink3}{cmyk}{0, 0.7808, 0.4429, 0.1412}
\definecolor{mygray}{gray}{0.6}
\usepackage{courier}

\newcommand{\gai}{generative artificial intelligence\,}
\newcommand{\Gai}{Generative artificial intelligence\,}

\newcommand{\eg}{e.\,g.,\,}

\newcommand{\cf}{{cf.\,}}

\def\BibTeX{{\rm B\kern-.05em{\sc i\kern-.025em b}\kern-.08em
    T\kern-.1667em\lower.7ex\hbox{E}\kern-.125emX}}

\title{Integrating Generative Artificial Intelligence in \\Intelligent Vehicle Systems} %

\author{Lukas Stappen$^{1}$, Jeremy Dillmann$^{1}$, Serena Striegel$^{1}$,\\ Hans-Jörg Vögel$^{1}$, Nicolas Flores-Herr$^{2}$, Björn W. Schuller$^{3}$ %
\thanks{$^{1}$BMW Group Research and Technology, 
        Munich, Germany
        {\tt\small stappen@ieee.org}}%
\thanks{
$^{2}$Fraunhofer Institute for Intelligent Analysis and Information Systems IAIS, Dresden, Germany
}
\thanks{
$^{3}$Imperial College London, London, UK and University of Augsburg, Augsburg, Germany and audEERING GmbH, Munich, Germany
}}

\begin{document}

\maketitle

\begin{abstract}
This paper aims to serve as a comprehensive guide for researchers and practitioners, offering insights into the current state, potential applications, and future research directions for generative artificial intelligence and foundation models within the context of intelligent vehicles. As the automotive industry progressively integrates AI, \gai technologies hold the potential to revolutionize user interactions, delivering more immersive, intuitive, and personalised in-car experiences. 
We provide an overview of current applications of \gai in the automotive domain, emphasizing speech, audio, vision, and multimodal interactions. We subsequently outline critical future research areas, including domain adaptability, alignment, multimodal integration and others, as well as, address the challenges and risks associated with ethics. By fostering collaboration and addressing these research areas, \gai can unlock its full potential, transforming the driving experience and shaping the future of intelligent vehicles. 
\end{abstract}

\section{Introduction \label{sec:intro}}
Generative artificial intelligence has witnessed an unprecedented growth, with applications such as ChatGPT becoming the fastest adopted consumer software in history~\cite{hu_2023}. These generative, general-purpose (language) models, characterized by their capabilities to creatively create text, audio~\cite{dhariwal2020jukebox}, and images~\cite{rombach2022high} in natural interaction, have swiftly permeated our daily lives. 
In the face of such rapid progress, it is essential to consider the impact of these \gai technologies on the future of intelligent vehicles and the transformative potential they hold for the automotive industry.

The vehicle presents a distinctive scenario, in which the driver is occupied with primary driving tasks, such as steering and accelerating, as well as secondary tasks like activating windshield wipers ~\cite{bubb2003fahrerassistenz}. \Gai will primarily enhance tertiary tasks, focusing on the control of in-vehicle infotainment systems, which contribute to the growing prominence of intelligent vehicles and the pursuit of realizing the vision of an all-round personal assistant within automobiles. BMW, for instance, envisions a digital companion~\cite{youtube_2023} that engages with users emotionally, cultivating a natural relationship through dialogue. Inputs and outputs across various modalities are vital to achieve the highest quality of interaction in high-stakes driving situations. This encompasses two components: streamlining necessary tasks, such as multi-turn point of interest searches for charging, parking, and recreation, and providing entertainment and productivity opportunities during unused driving periods. Additionally, several hedonic elements are crucial in crafting an emotional experience within the vehicle~\cite{vogel2018emotion}, including in-car visualization (\eg lighting), audio functions (\eg e-vehicle sounds), and exterior features (\eg light carpet). The creative prowess of \gai offers the ideal tools for actualizing this vision and forging the future of intelligent vehicles.

The spectrum of \gai approaches is mostly bounded to training types, each with its unique advantages in specific situations, as shown in \Cref{table:genai_techniques}. Unsupervised learning employs models like generative adversarial networks (GANs) and variational autoencoders (VAE) to generate data samples. The StableDiffusion architecture, for instance, combines the latter with a reconstruction (diffusion) process to create photo-realistic images~\cite{rombach2022high}. Besides images, also audio can be generated without requiring labelled data, making it valuable when labels are scarce~\cite{wang2018high}.
Supervised generative modelling uses architectures like conditional GANs (\eg pix2pix), which are trained with a large amount of labelled data, allowing them to generate data samples conditioned on specific attributes and making them most useful when data must adhere to certain conditions~\cite{isola2017image}. The 
training technique of most foundation models, models adaptable to a variety of down-stream tasks, are usually trained with self-supervised learning, such as those employed by autoregressive transformer models like GPT-3~\cite{floridi2020gpt}, learn by creating their own supervision signal from the input data and thereby able to utilise vast quantities of data. Hereby, they are predicting parts of the input from the rest of the input, such as the next word in a sentence, making them advantageous when seeking to generate coherent and contextually relevant output~\cite{floridi2020gpt}. Lastly, reinforcement learning became recently more popular to learn complex behaviours, for instance, fine-tuning the model to align with human expectations using proximal policy optimization~\cite{ouyang2022training}. By first learning a supervised reward function, which then acts as a judge for the generated data samples, the model outputs can be sharpened automatically within a closed-loop reward system.

\begin{table*}[t]
\centering
\resizebox{.75\textwidth}{!}{%
\begin{tabular}{c|c|c|p{7cm}}
\hline
\hline
\textbf{Training} & \textbf{Generative model} & \textbf{Domain} & \textbf{Intelligent vehicle applications} \\ \hline
\hline

\multirow[]{6}{*}{Unsupervised} & Jukebox \cite{dhariwal2020jukebox} & \makecell{Audio\\ Music Generation} & Generates music, including raw audio, in various styles and genres, \eg entry soundtrack composition \\ 
& StyleGAN2~\cite{karras2020analyzing} & \makecell{Vision\\Image Generation} & Generates high-resolution, photorealistic images, such as faces and landscapes \eg visual assistant avatar \\ %
& StableDiffusion~\cite{rombach2022high} & \makecell{Multimodal\\Image Generation} & Combines VAE with a reconstruction process to create photo-realistic images \eg exterior LED projections \\ %
\hline 
\multirow[]{12}{*}{\makecell{(Self-)\\Supervised}} 
& GPT-3/-4~\cite{openai2023gpt4} & \makecell{Text \\ Language Generation} & Generates human-like text based on input and context \eg natural assistant interaction \\ %
& RETRO~\cite{borgeaud2022improving} & \makecell{Text \\ Language Generation} & Improves language models by retrieving from trillions of tokens \eg intuitive retrieval from manual via Q\&A \\ 
& Tacotron 2~\cite{shen2018natural} & \makecell{Speech\\Text-to-Speech} & Generates high-quality, natural-sounding speech from text \eg assistant voice synthesis \\ %
& \makecell{MusicLM~\cite{agostinelli2023musiclm}} & \makecell{Audio \\ Music Generation} & Generates music with a coherent structure and long-term dependencies \eg EV sound modelling \\ %
& pix2pix~\cite{isola2017image} & \makecell{Vision\\ Image-to-Image } & Converting images from one domain to another such as sketches to photos \eg passenger entertainment \\ %
& DALL-E~\cite{mishkin2022risks} & \makecell{Mulitmodal\\Image Generation} & Generates images from textual descriptions \eg theme based lighting \\ %
\hline
\end{tabular}%
}
\caption{Excerpt of the latest generative models categorised by training technique and domain including potential intelligent vehicle applications.}
\vspace{-1.1em}
\label{table:genai_techniques}
\end{table*}

AI-enhanced vehicle functions presumably play a vital role in the mobility of the future and can be broadly categorized into perception and generation systems. Perception models are extensively researched for understanding interactions within vehicles~\cite{xaware2020, stappen2020domain} and anticipating passengers' behaviour and emotional states~\cite{zepf2020driver}. 
Research involving generative models is considerably less substantial. They have been employed to enhance visual classification using VAEs for unsupervised domain adaptation between automotive sensors~\cite{reis202}. Similarly, the authors of ~\cite{wang2018high} utilized conditional GANs to improve the capabilities of autonomous vehicles by simulating more realistic and varied real-world scenarios.
In the realm of audio generation, SoundsRide~\cite{Mohamed2021} introduces an in-car audio system that synchronizes music with sound affordances along the ride in real-time, providing an immersive music experience that could potentially influence driving safety, with both positive and negative effects depending on the mix and user.
Improving dialogue and personalised voice characters for in-car speech interfaces,~\cite{braun2019} examines the design and impact, comparing four assistant personalities (friend, admirer, aunt, and butler). The findings suggest that matching the user's personality results in higher likability and trust. However, existing generation research is limited in terms of breadth and depth of the applications, as well as in multimodal modelling.

This paper adopts a practical approach, seeking to develop a research agenda that explores the application of \gai technologies in intelligent vehicles. We examine the role of AI in augmenting intelligent vehicle functions and the potential of \gai to facilitate multimodal interaction, encompassing audio, video, and speech in these systems. Our research agenda is guided by key principles, such as model capabilities, ethical considerations, and the alignment of AI models. Furthermore, by presenting specific use cases along the modalities, including productivity and relationship-building functions, we lay the foundation for research challenges and opportunities associated with these applications, paving the way for future research directions in this burgeoning field.

The paper is organised as follows:
~\Cref{sec:applications} explores the role of \gai use cases from various modalities. ~\Cref{sec:principles} delves into the principles guiding our research agenda and discusses the challenges and opportunities presented by the integration of \gai in intelligent vehicles. ~\Cref{sec:implications} covers implications such as an interdisciplinary approach to collaboration and standardization, as well as an examination of potential risks and system limitations. Overall, our research agenda endeavours to advance the understanding and application of \gai technologies in intelligent vehicles, with far-reaching implications for safety, user experience, and industry innovation. By casting a discerning eye on the potential of these technologies and their impact on the automotive landscape, we hope to contribute to the discourse surrounding the future of intelligent vehicles and the exciting possibilities that lie ahead.

\section{Potential Applications of generative AI in intelligent vehicles  \label{sec:applications}}
In this section, we explore various use-case ideas for \gai in intelligent vehicles, categorized across the three modalities of speech, audio, and video (\cf \Cref{table:genai_techniques}). 
Hereby, the focus is primarily on human-machine interaction.
As \gai holds the potential to significantly enhance various aspects of intelligent vehicles, it is irremissible that future perception systems accurately identify and target specific interactors, thereby preventing confusion for those who are not the intended recipients of the interaction. 

\subsection{Speech \label{sec:voice}}
\begin{figure}[!t]
\centering
\includegraphics[width=\linewidth]{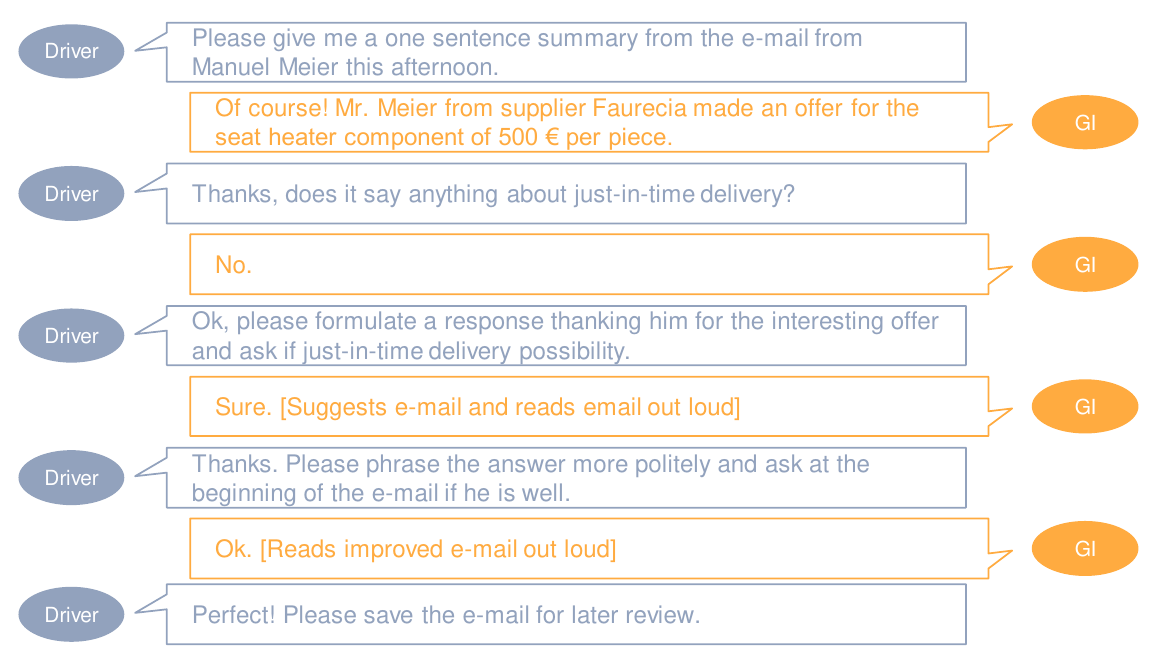}
\caption{Example communication flow between a driver and the \gai personal assistant, reading and answering emails in a productivity in-car suite, conducting \gai tasks, such as summary, interactive formulation, and searching and question \& answering. }
\label{fig:productivity}
\vspace{-1.1em}
\end{figure}

Generative voice AI can greatly enhance the voice capabilities of intelligent vehicles, offering users a more engaging experience. Voice interaction is an ideal modality for drivers to search for information and receive decision support, \eg which charging station to use, which road to choose, without having to take their eyes off the road. This modality can offer both user-experience and safety benefits. However, voice interaction is only engaged if the system does not disappoint drivers and trust is built. By facilitating more humanlike dialogues~\cite{openai2023gpt4}, generative models based in vehicle-assistant can provide context-aware and personalised responses to queries and requests, making interactions seamless and efficient. Furthermore, a productivity assistant (\cf ~\Cref{fig:productivity}) can help users draft and edit emails, prepare presentations, or perform other tasks through enhanced voice interactions, contributing to a more productive in-car environment as captured in.
Emotional voice~\cite{shen2018natural} and personality can also be generated by \gai systems, allowing the voice assistant to express emotions and adapt its tone to match the brand, user preferences, or specific scenarios. This personalisation fosters a stronger connection between the user and the vehicle. Additionally, \gai can provide real-time translation services~\cite{radford2022robust} for radio traffic warnings or conversations between driver and passenger, breaking down language barriers and aiding communication during foreign travel or with passengers speaking different languages.

\subsection{Audio \label{sec:audio}}
Generative audio AI possesses the capacity to revolutionize the auditory experience, providing users with an unparalleled level of personalization and immersion. By generating bespoke welcome melodies upon the user's entry, vehicles can evoke a sense of familiarity and belonging~\cite{payne2019musenet, dhariwal2020jukebox}. With the increase of semi-autonomous vehicle functions and thus regular driver engagement by the vehicle, a personal set of auditory warning signals as well as a guidance system for people with disabilities in fully-autonomous driving are possible. Furthermore, electric vehicles can benefit from AI-generated in-car sound modelling that simulates engine noises or other bespoke audio experiences~\cite{agostinelli2023musiclm} for passengers, adhering to safety concerns by not affecting exterior noise levels. 

\subsection{Vision \label{sec:vision}}
The realm of vision presents yet another avenue, offering a myriad of visually engaging and interactive features. AI-generated persona avatars~\cite{karras2020analyzing}, for instance, can actively listen, empathize, and respond with emotions, fostering a more profound connection between the user and the vehicle's assistant. The appearance can also individualise projecting visuals to intended recipient screens, \eg the driver head-up display. This enriched interaction transcends the limitations of conventional voice-only communication. Furthermore, the navigation and information system can leverage \gai to adapt the visual appearance~\cite{mishkin2022risks} context-aware to ambient light sensors, weather data, user preferences, interior design and the type of road. In addition, it can create custom album art, artist portraits, or visualizations based on the audio being played. AI-generated visual content~\cite{rombach2022high} can also be employed to create personalised LED night sky displays and interactive animations projected from the door handle. These personalised touches can elevate the in-car entertainment experience adding an element of wonder and fascination. Moreover, \gai can assist in creating visual summaries of accident reports, which can be relayed to emergency services, streamlining the process and potentially improving response times.  
As shown in~\cite{Zijie2020}, \gai hold also promises for enhancing training data for driver assistance systems by creating realistic, diverse, and high-quality synthetic data. Also, a visual style transfer from one country's signs to another in the form of typical colours and symbols is possible. As mentioned before, however, the practical implementation of such models in these safety-critical domains can be challenging, due to the stringent regulatory environment.

\subsection{Multimodal \label{sec:multimodalApp}}
The ongoing evolution of \gai systems is characterized by a shift toward multimodal models~\cite{villegas2022phenaki}, capable of emulating human cognition by seamlessly processing and integrating multiple modalities. This comprehensive approach allows for a more profound understanding of user needs, fostering a richer, more intuitive user experience~\cite{xaware2020}. 
Take, for instance, vehicle issue diagnosis: A multimodal AI system could astutely comprehend a user's verbal description of a problem while concurrently analysing visual cues from the vehicle's sensors or user-provided images. This synergistic processing of information potentially results in a faster, more precise diagnosis and customized suggestions. Furthermore, when elucidating vehicle functions, the AI assistant can employ a harmonious blend of voice and visual modalities to offer lucid, methodical guidance, thereby enhancing user comprehension and the overall driving experience. By embracing the sophistication of multimodal \gai, intelligent vehicle systems are poised to revolutionize user interactions, ushering in a new era of captivating and instinctive in-car experiences.

\section{Key Principles Guiding Future Research on generative AI in Intelligent Vehicle Systems \label{sec:principles}}
In this section, we argue that the following principles are fundamental to the advance and implementation of \gai 
for intelligent vehicles: 

\begin{itemize}
\item Domain adaptability and personalisation
\item Reliability, alignment, and controllability
\item Multimodal integration
\item End-to-end architecture
\item Ethical considerations
\end{itemize}
These principles serve as a foundation for successful design and implementation of innovative in-vehicle functions that are both efficacious and user-centric. In the following, we describe these principles in detail and show how they are instrumental in identifying crucial research topics and potential avenues in the domain of \gai in intelligent vehicles.

\subsection{Domain Adaptability and Personalisation \label{sec:personalisation}}
\Gai systems should be designed to adapt to specific domains, integrating non-public knowledge in the model. In addition, models should also adapt to user behaviour, preferences, and contexts to deliver personalised experiences. This principle ensures that the AI algorithms can tailor their interactions and responses to suit individual industries and users, leading to greater satisfaction and engagement. In practice, this might involve generating customized routing explanations based on the driver's preferred scenic routes or dynamically adjusting the lighting in-car environment to suit their mood and preferences.

A deciding research direction lies in the adaption of general-purpose, generative models to the automotive domain, especially for context-driven language models. From a practical standpoint, automotive companies are either provided with injecting domain knowledge into the model of providers or find themselves compelled to train their own general-purpose models. This domain knowledge is often not publicly accessible and might change rapidly, therefore, existing prompt-based fine-tuning methods are not sufficient. One direction could be retrieval transformers, incorporating external knowledge sources and long-term context by a retrieval mechanism~\cite{borgeaud2022improving} while inadvertently staying further aligned to the overarching vision of artificial general intelligence. 

Personalisation requires to cater for individual user preferences accurately and hold limited context over a long time. For example, language models have experienced a significant increase in token capacity, expanding from a mere 2k tokens to a seemingly ample 32k tokens with GPT-4~\cite{openai2023gpt4}. When considering the vast scope of user interactions that span hours, weeks, or even months, this limitation appears still rather restrictive. Chaining representation of context in the form of embeddings as provided by open-source libraries\footnote{https://github.com/hwchase17/langchain} could be one direction. To preserve control over the model, major non-open-source providers do yet not offer such services. 
In speech personalisation, synthesis still suffers from unclear and missing words that can be attributed to disordered attention alignments in the phoneme-to-acoustic part of autoregressive models~\cite{wang2023neural}. This, combined with a lack of diversity in training copra, results currently in shortcomings to accurately representing key aspects of speaking styles such as accents and prosody and should be addressed in the near future. Tailoring music and visualisations to individuals using foundation models is still in its infancy. A promising starting point involves the expending of the source and representation of input data beyond the conventional text inputs. This may include incorporating sensor data or time series of events that implicitly capture changes in human behaviour and potentially achieve a deeper, more automatic generation. 
Therefore, central research questions remain open, such as what approaches can be employed to develop AI systems that can dynamically adapt to an individual user context that will steer future investigations toward the development of highly domain adaptable and personalised \gai solutions.

\begin{figure}[t]
\centering
\includegraphics[width=\linewidth,trim={.2cm .5cm 1.cm .8cm},clip]{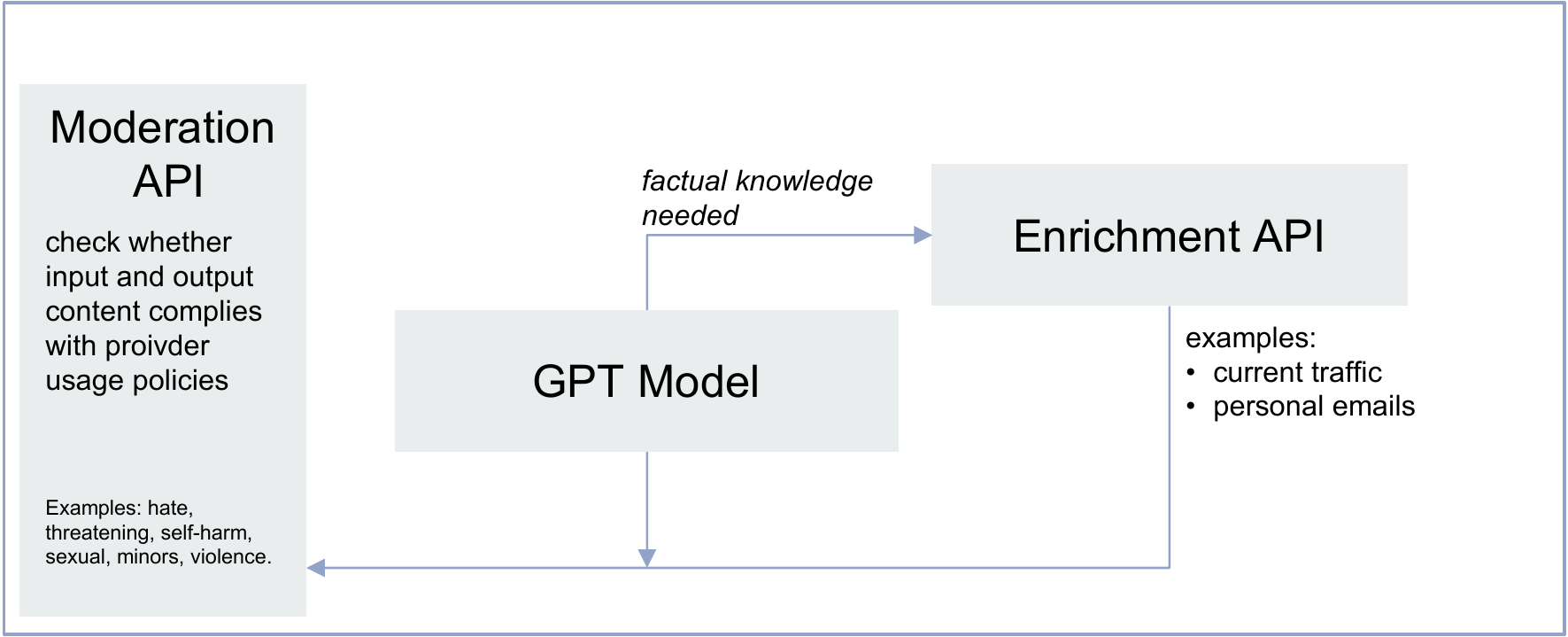}
\caption{A schematic illustration of a controllable \gai voice user interface. The user prompt is firstly moderated, given to the general-purpose model to secondly either be directly answered, or thirdly, optionally enriched by non-model data using external API calls, before the natural answer is lastly given back to the user via a moderation layer.}
\label{fig:GPT_Model}
\vspace{-1.1em}
\end{figure}

\subsection{Reliability, Alignment, and Controllability \label{sec:alignment}}
Ensuring that \gai models for intelligent vehicles are reliable, aligned, and controllable is crucial. Reliability encompasses the model's consistent performance across diverse inputs resulting in trustworthy outputs, while alignment refers to adherence to human as well as company values and standards. Controllability enables fine-grained control over the model's output, ensuring generated content is contextually relevant and brand-aligned.
By addressing these concerns, \gai models can be developed that meet user needs and preserve brand integrity, while mitigating risks associated with inappropriate or misaligned content. This principle fosters trust in AI systems integrated into intelligent vehicles.

This pursuit demands the exploration of effective fine-tuning strategies, encompassing both broad domains and specific use-cases, to steer the system reliably toward their designers' intended goals and interests and minimize the risk of generating inappropriate content. First, it is crucial to develop strategies for identifying appropriate internal and external sources to align the model~\cite{nakano2021webgpt}. By meticulously training with well-selected data samples, the model should minimize the risk of generating hallucinations, which are seemingly coherent but ultimately nonsensical or incorrect outputs during inference. Second, despite recent improvements in local (follow user instructions) and global (follow tone rules) language model alignment, it is questionable if the inherent nature of an autoregressive model will ever enable sufficient alignment~\cite{openai2023gpt4}. Practically speaking, for controlling that the generation stays in each operator's own responsibility and independent, robust moderation mechanisms (\cf \Cref{fig:GPT_Model}) are imperative. %
Thus, the guiding research question is how \gai systems can be made more reliable to ensure their outputs align with user intentions, company values, and safety requirements, particularly when dealing with uncontrolled inputs and outputs.

Another pivotal research area involves the controllability of how the model orchestrates the output. Long-lasting information can be stored and retrieved within the model itself. Short-lived information requires real-time system data, such as traffic, for efficient route comparison. Thus, models must effectively learn when, which, and with what arguments an external API should be called, and how to integrate this information controllably into the output (\cf \Cref{fig:GPT_Model}). First approaches such as the Tooltransformer~\cite{mialon2023augmented} might be a first step into a full generic or automotive-specific direction. A component of this research may involve transitioning between legacy and \gai systems. Particularly, in the primary application of personal assistant systems, the shift towards a fully developed \gai system will occur progressively. Future studies must explore hybrid systems that enable seamless transitions between responses generated from both \gai and static, legacy system components.

\subsection{Multimodal Integration \label{sec:multimodal}}
A paramount principle in the development of \gai for intelligent vehicles is the seamless integration of different modalities, such as speech, audio, and vision. This integration fosters context-aware and natural interactions, enhancing the user experience. For example, a holistic approach to understanding driver commands and gestures can lead to improved responses from the vehicle's assistant, creating a more intuitive experience.

The pursuit of advanced multimodal integration in \gai systems for intelligent vehicles presents a multifaceted and intriguing avenue for future research. Moving beyond conventional image-to-text pairings as present in current model developments~\cite{rombach2022high}, scholars and practitioners alike must delve into the intricacies of time-continuous vehicular scenes, investigating practical sampling rates for sensing and exploring methodologies for feeding inputs from multiple modalities into models. By addressing such inquiries, researchers will gain insights into the precise circumstances under which reliance on a single modality may be sufficient, as opposed to scenarios where multimodal integration is indispensable for a thorough understanding of user needs and context. Consequently, the overarching research question -- how can \gai models be designed and trained to effectively integrate and process information across multiple modalities, such as speech, audio, and vision -- shall propel the field toward the development of sophisticated, context-aware systems that enhance the experience.

\subsection{End-to-End Architecture \label{sec:tuning}}
\begin{figure*}[t]
\centering
\includegraphics[width=.7\linewidth]{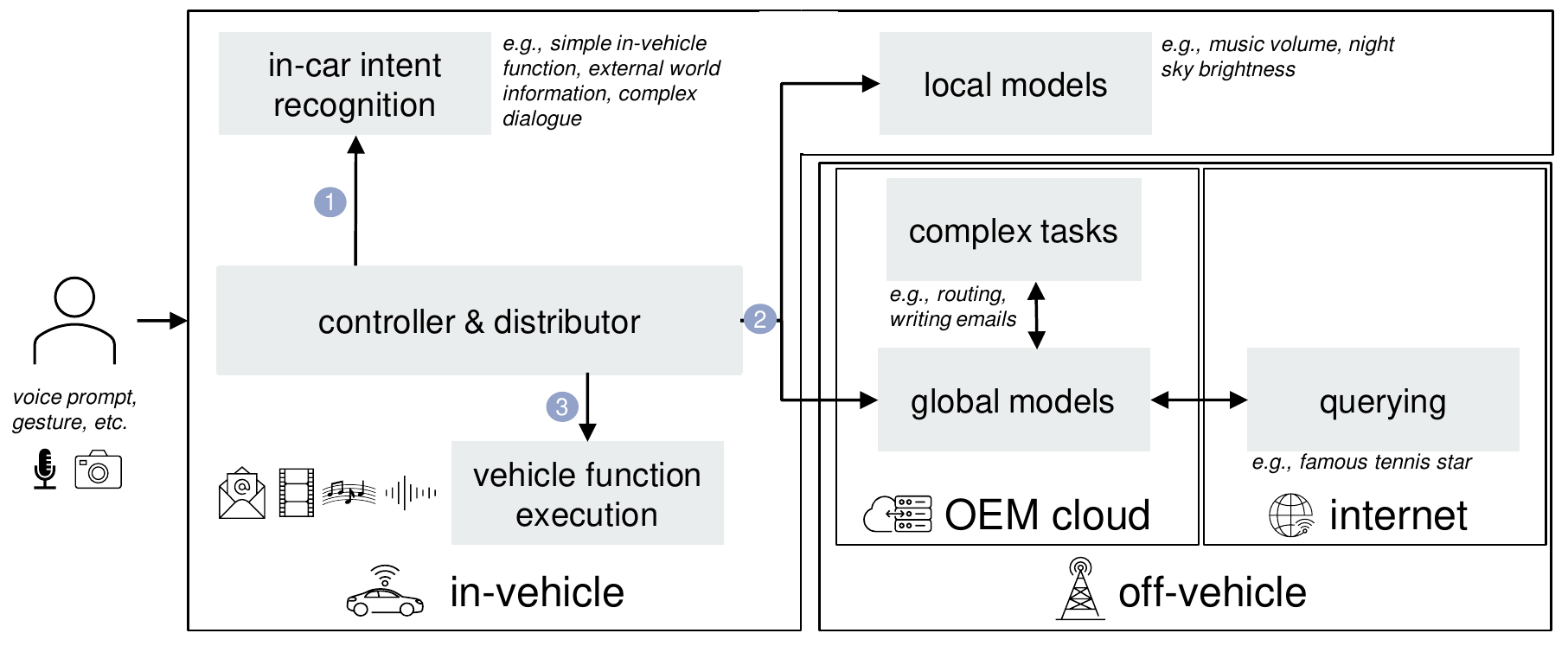}
\caption{Multimodal flow from user interaction, hybrid computation distribution depending on task complexity and considering the type of model, to the execution of a vehicle function. }
\label{fig:ipa_flow}
\vspace{-1.1em}
\end{figure*}

While many existing use-cases demonstrated by prominent AI companies operate in API-driven software platforms with no hardware involvement, the nature of vehicular systems necessitates a more intricate principle. The fusion of hardware and software components calls for end-to-end thinking, from the user interface, computation separation (hardware) and model sizing (software) as illustrated in \Cref{fig:ipa_flow}.

\subsubsection{User interface}
In spite of the potential to facilitate an unprecedented user-experience, \gai solutions will only be adopted if the multimodal systems are able to gain users'  trust. This is especially true in-vehicle \gai use-cases that are performed next to the driving task. To develop trust, the classical user-interface design guidelines such as transparency, consistency, controllability, and easy error correction may not be sufficed ~\cite{amershi2014power}. User interface solutions need to take into account a degree of uncertainty present when interacting with \gai systems ~\cite{sarkar2015confidence}. For example, research is necessary to understand if and how drivers recognize and manage potential faulty predictions by the \gai system ~\cite{bucher2017algorithmic}. The technical improvement of \gai solutions must be accompanied by human-computer interaction research, involving iterative prototyping and user testing ~\cite{eiband2018bringing}. Future \gai interfaces must allow users to acquire and apply knowledge on how they can interact with their \gai system to achieve use-case specific goals ~\cite{eiband2018bringing}.

\subsubsection{Distributed computation}
Another decisive research direction is the balance between in-vehicle and off-vehicle computation. The choice between an edge, hybrid, or cloud architecture for \gai applications in intelligent vehicles can be influenced by available hardware, the complexity of the task, and the desired performance. The later two are strongly linked to the size of the model (\cf~\Cref{sec:model}). Tasks may range from shallow software and hardware integration (\cf night sky~\Cref{sec:vision}), to deep integration such as intent routing within existing personal assistants. An example of deep integration is depicted in~\Cref{fig:ipa_flow}, where intent recognition directs user prompts between in-vehicle recognition and executions (\eg adjusting music volume), external world information (\eg details about a tennis star), and automotive use-cases (\eg complex route comparing) solved by \gai that are also done off-vehicle due to a hybrid or legacy architecture. We expect that such systems will form the transition until all parts can be reliably taken over by \gai systems. 
Furthermore, as \gai models evolve rapidly, the practicality of over-the-air updates must be weighed against the computational demands of model inference and the highly specialized nature of current hardware. %
Critical aspects of an overall architectural design and integration are ongoing research matters. 

\subsubsection{Model type and size\label{sec:model}}
Closely connected to where the computation takes place is the type and size of the model. \Gai models are generally seen as large, therefore, resource-intensive compared to other machine learning models. For instance, smaller models may be employed for relatively simple tasks, such as generating personalised welcome melodies. These models may offer faster response times and lower computational requirements, making them suitable for in-vehicle integration (\cf ~\Cref{fig:ipa_flow}). Conversely, more complex tasks, such as reasoning over the internet, real-time and historical data when creating natural language or detailed visual content, may necessitate larger models with billions of parameters, computed in the cloud, that can capture the nuances and intricacies of human language or visual elements.
Researchers must strive to explore novel techniques for model pruning since currently \gai models are undertrained, thus, inefficient~\cite{hoffmann2022training}. Furthermore, fine-tuning that simultaneously promotes low-latency natural interactions and maximizes the efficiency of model architectures is key. It is unclear yet how automotive-specific \gai models can be developed and optimized to reconcile these competing demands. Currently, they also have a significant influence on the training and inference costs and are contrary to green AI initiatives. 
As the field of \gai advances, the prospect of devising a unified model that adeptly accommodates a vast array of automotive use-cases emerges as a compelling research direction. Such a versatile model would necessitate striking an optimal balance between performance, computational requirements, and resource constraints, especially within the realm of embedded systems and real-time applications.

Future investigations should focus on optimizing integrated architectures, model size, and computational efficiency to the creation of streamlined AI systems that can adeptly navigate the multifarious challenges and complexities inherent to the modern automotive landscape.

\subsection{Ethical Considerations \label{sec:ethics}}
As \gai will become increasingly integrated into intelligent vehicles, it is pivotal to address ethical concerns that may arise, including privacy, security, and fairness. Transparent data collection and processing practices, as well as rigorous security measures, must be in place to safeguard user data and ensure trust. Those principles are not new in the area of AI, however, remarkably still treated with negligence resulting in a complete ban of ChatGPT in Italy and ongoing debates in Germany, arising from unaddressed concerns surrounding training and user data.

It is essential to develop \gai models that are unbiased and equitable, fostering fair and inclusive in-vehicle experiences for all users. For instance, when imitating voice profiles, the system should be capable of accurately simulating various accents and dialects without perpetuating stereotypes or biases. By doing so, AI-generated voices can provide a more diverse and inclusive representation. For text generation, limiting harmful and wrong content is crucial. Challenges arise from the limited understanding of model learning, typically addressed by explainable AI. Since the predictions in generative models are not discrete labels but rather generated artifacts, interpreting them becomes difficult due to their high degree of variability. Atman presents an intriguing direction by employing perturbations to alter the attention mechanisms, enabling the suppression or emphasis of specific tokens within the transformer model~\cite{deb2023atman}. By exploring alternative generations synthesized through this manipulation, researchers can gain a better understanding of how foundation models generate content, potentially leading to more controlled and adaptive content generation.

Navigating the intricate landscape of ethical considerations in the development and integration of \gai systems for automotive applications necessitates meticulous collaboration between researchers and policymakers. Establishing clear ethical standards and guidelines will be instrumental in directing the responsible development and deployment of these technologies. Drawing from the EU AI Act~\cite{veale2021demystifying} which might provide a basis for other countries in developing their regulatory frameworks related to AI in the future, several pertinent research questions emerge: How can AI systems be accurately classified, and the potential risks associated with \gai systems in the automotive sector be assessed? How do we ensure that \gai systems interacting with humans, generating content, or recognizing emotions align with stringent ethical categories and comply with legal requirements regarding transparency, accountability, data quality, documentation, and human oversight? What practical approaches can be devised to store and manage sensitive data, striking a balance between external service providers and original equipment manufacturers, while adhering to privacy regulations and user preferences?
Considering the planned transitional phase of 24 to 36 months before the EU AI Act starting from Q2 2023 comes into full effect, these research questions, among others, will have to be addressed swiftly to shape future inquiries, ultimately guiding the ethical development and integration of \gai technologies in the automotive domain.

\section{Implications for Generative AI Research \label{sec:implications}}
The principles outlined emphasise the need for an interdisciplinary approach, fostering collaboration and addressing challenges in \gai technologies for intelligent vehicles. This involves cooperation among researchers, industry stakeholders, and academia to establish best practices, benchmarks, and shared resources, while also exploring potential risks and system limitations. %

\subsection{Collaboration and Standardisation \label{sec:dirCollaboration}}
Given that the research landscape and goals (\eg alignment~\cite{ouyang2022training}) are discovered and heavily influenced by corporate entities, such as OpenAI and Microsoft, a collaboration between AGI players, domain adopters, and research institutions is crucial to identify areas where collaboration can be bolstered. Furthermore, training data, open-source models, and the establishment of best practices and benchmarks are integral components for fostering a shared understanding and facilitating advancements. Consequently, effective collaboration between researchers and industry stakeholders is essential to establish best practices, benchmarks, and shared resources for the development, evaluation, and deployment of \gai technologies in intelligent vehicles, while balancing the interests and contributions of both corporate and academic entities.

\subsection{Potential Risks and System Limitations \label{sec:risKAndLimitations}}
It is essential to recognize and address potential risks and limitations that may not be entirely mitigated, ultimately transforming our lives and the way we interact with the world.
Although, if \gai models are thoroughly tested and validated to become safe to operate, free user input is prone to adversarial attacks, which likely can never be fully mitigated by countermeasures. Implementing stringent privacy and data protection measures is vital to maintain user trust and adhere to regulatory requirements. For example, the liability to cyberattacks can be reduced, but never completely ruled out, making an exposure of highly private data on an immense scale. Developing guidelines to address system breaches, not only involving individual operators but also entire systems, such as nations, could be a valuable endeavour.

Even though they are virtual systems, such systems directly influence the physical world when people act on their suggestions. As trust in their reliability grows, so does the blind dependence on these systems, especially since humans are often unaware of black swan events. The combination of overconfidence (as seen in hallucinations) and concurrent sensory gaps may result in life-threatening consequences. A straightforward example in the automotive industry could involve a user being inadvertently guided into a hazardous situation due to a faulty sensor falsely indicating a cold engine, prompting the system to suggest an oil change on a hot engine.

\section{Conclusion \label{sec:conclusion}}
The advent of \gai technologies has opened up a plethora of opportunities and challenges for the automotive industry, paving the way for the development of intelligent vehicles that provide more immersive, intuitive, and personalised in-car experiences. This paper has presented an overview of current applications and future research directions in the domain of \gai and intelligent vehicles, highlighting the potential of these technologies to revolutionize user interactions and drive innovation in the sector.
Key future research areas in \gai and intelligent vehicles include multimodal integration, model optimization, personalisation, reliability, and architecture. Additionally, ethical implications particularly focusing on user privacy, data security, and potential misuse are of importance. Addressing these areas by fostering collaboration will unlock \gai´s full potential, transforming the driving experience and shaping the future of intelligent vehicles. 
One limitation of the work is that we have maintained a perspective where driving remains the primary task of the driver. While many of the scenarios discussed may still be relevant in a fully automated context, it is important to acknowledge that a shift in priorities could occur. For instance, the interaction modality might transition from a focus on voice towards a greater emphasis on graphic user interfaces, and use-cases designed to combat boredom (passive fatigue) might see an increase in demand.
Nonetheless, as \gai continues to advance, the development of intelligent vehicles that cater to the diverse needs and preferences of users will become increasingly important, ultimately redefining the way we interact with and experience the world around us.

\bibliographystyle{unsrt}
\bibliography{refs}

\end{document}